# The Acquisition of a Lexicon from Paired Phoneme Sequences and Semantic Representations

Carl de Marcken
MIT Artificial Intelligence Laboratory
545 Technology Square
Cambridge, MA 02139, USA
cgdemarc@ai.mit.edu

**Abstract**

We present an algorithm that acquires words (pairings of phonological forms and semantic representations) from larger utterances of unsegmented phoneme sequences and semantic representations. The algorithm maintains from utterance to utterance only a single coherent dictionary, and learns in the presence of homonymy, synonymy, and noise. Test results over a corpus of utterances generated from the Childes database of mother-child interactions are presented.

## 1 Introduction

This paper is concerned with the machine-learning of a lexicon from utterances that consist of an unsegmented phoneme sequence paired with a semantic representation of what those phonemes collectively mean. The problem is modeled after the environment that a child learns in, presented with a continuous speech signal[1] and potentially hypothesizing a meaning for that signal based upon visual stimuli. We radically simplify the problem the child encounters for the computer by pre-digesting the speech stream into a sequence of phonemes, and by providing an exact, transparent, and unambiguous semantic representation.

For instance, the computer might be presented with the utterance[2] "OK, here's the big ball":

| Phoneme Sequence | Sememe Set |
|---|---|
| /ɑkhirzðəbɪgbɔl/ | { BE THE OK BALL HERE BIG } |

---

[1] Children do in fact hear some pause information, and thus restricting our algorithm to totally unsegmented speech is somewhat unnatural. But, as data in [1] suggests, short utterances often are pauseless, and many sentences children hear are quite short (5.6 words on average in our test database, with little embedding).

[2] The particular phonemes used in the paper are the output of a public domain text-to-phoneme converter, which is frequently inaccurate (witness "OK" → /ɑk/). It's mislabelings do not have any effect on the work presented here.

From this and other utterances our goal is to produce an algorithm that learns a lexicon containing:

| /ɑk/ | { OK } | /hir/ | { HERE } | /z/ | { BE } |
|------|--------|-------|----------|-----|--------|
| /ðə/ | { THE } | /bɪg/ | { BIG } | /bɔl/ | { BALL } |

It is not sufficient for our algorithm to learn any mapping from phoneme sequences to semantic symbols that explains the training data. Many mappings will do so, including the trivial one in which the final lexicon contains the training utterances themselves. We ask that the algorithm learn a lexicon that generalizes well, and does so without recourse to cognitively implausible mechanisms, such as off-line algorithms that access large amounts of the corpus at once.

This problem is interesting for several reasons. Although considerable study has been devoted to the acquisition of formal and natural grammars, grammatical categories for words, and even phonological processes, the acquisition of the lexicon has been largely neglected by the computational linguistics and machine learning communities, despite growing agreement that most language variation stems from there. Thus from a cognitive science and artificial intelligence viewpoint, this problem is a fundamental but relatively unstudied part of the process of learning a natural language, and a prerequisite to the acquisition of grammar. It also is a task where model complexity must be traded against global coverage in an environment where only a small proportion of the data is available at any one time to the algorithm.

Of course, the work presented here presumes a gross simplification of the real task faced by a computer or child that seeks to learn from sound pressure waveforms and sensory stimuli. Children may well not have the innate capability to segment a speech stream into discrete alphabetic sound units like phonemes (this is a matter of some debate in the field) and even if they do, word pronunciation is highly context-dependent. This is not necessarily an insurmountable problem; the methods used here can easily be extended to handle some sound variation in words, and section 6 discusses more recent work that accepts input closer to what current technology could derive from sound waves. A bigger assumption is the uniqueness and simplicity of the semantic representation for each utterance; the simplicity of our semantic representation both complicates the learning process for our algorithm by eliminating information that could constrain the search (in conjunction with grammar, for example) and simplifies the task by reducing the complexity of the information to be learned. The unambiguous interpretation we implicitly assume the child can give an utterance in an environment may or may not be trivializing the task: given our limited knowledge of child psychology it is difficult to tell. But some methods for increasing the ambiguity of semantic interpretation for an utterance are discussed in the next section and in section 6.

## 2 Previous Work

Olivier [9] and Cartwright and Brent [4] present simple algorithms that segment text and phoneme sequences by learning from statistical irregularities. In particular, they place phoneme sequences in a dictionary if those phonemes occur consecutively more often than they would if phonemes were selected by a memoryless random process with identical aggregate distribution. Olivier's algorithm is on-line, extremely efficient, and incorporates no priors. Cartwright and Brent's is a batch description-length formulation [10] that uses the size of the dictionary as a prior. Their algorithms perform with minimal adequacy, unable to distinguish correlations due to the dictionary from correlations due to syntax and semantics.

Brown *et al* [2] present a statistical machine translation algorithm that makes use of estimated correspondances between words in English and French. From an aligned multilingual database they estimate for every English word the distribution of French words it might translate to, including the number of words it will translate to. So, they estimate that *not* will translate to *pas* (.469), *ne* (.460), *jamais* (.002) ..., and that it translates into 2 words with probability .758 (*ne pas*). If we were to segment our phonemic input into words we could view our problem as acquiring translation data from the language of sound to the language of meaning (or vice-versa). Brown *et al* of course assume segmentation of the source and target language, and make multiple passes over the data. They also make use of generally linear correspondances between utterances in the two languages.

Siskind has presented a series of algorithms [11], [12], [13] that learn word-meaning associations when presented with paired sequences of tokens and semantic representations. Our model of our problem is based on his work. His work differs from ours in two principle ways: first, Siskind learns more complex semantic representations (Jackendoff-style semantic formulae [7] rather than simple symbol sets[3]), in an environment where his algorithms are presented with many ambiguous semantic representations (one of which is correct); and second, Siskind's work assumes pre-segmented tokens as input. So, his algorithm receives *ok here is the big ball* rather than our /ɑkhirzðəbɪgbɔl/. Siskind's work has tended to rely on classical search methods, and maintains a dictionary that may contain a variety of concurrent hypotheses for any given word.

See the above papers for references to other related work, and a further discussion of the motivation for this line of research.

## 3 The Algorithm

The algorithm presented here maintains a single dictionary, a list of words. Each word is a triple of a phoneme-sequence, a set of sememes (semantic symbols), and a confidence factor called the temperature. The temperature affects the likelihood that the word will be spontaneously deleted from the dictionary and also the ability of the word to participate in the creation of new words. Thus a high temperature (near 1) implies that a word is likely to be involved with new word creation and to be deleted, and a low temperature (near 0) implies that the word is static and unlikely to be deleted.

---

[3]In recent work Siskind has separated the learning of the set of semantic primitives associated with a word from the learning of the relations between those primitives. Borrowing from his work, it would not be difficult to extend our algorithm to learn semantic formulae rather than merely sets of semantic primitives. Similarly, the method Siskind uses to disambiguate between multiple ambiguous semantic interpretations for an utterance is equally applicable here. He essentially uses Bayes theorem to calculate the probability of a meaning given a word sequence from the probability of the word sequence given the meaning, which can be very effective if the proper definitions of some of the words in the utterance are known.

When the algorithm is presented with an utterance, it performs a local variation of the expectation-maximization (EM) procedure [6]: it attempts to parse the utterance using the words in its dictionary, resulting in values for hidden word-activation variables. By parse we mean that the algorithm attempts to find a set of words that collectively cover all the phonemes and sememes of the utterance, without overlap or mismatched elements. After the parse is complete, the maximization step occurs, with modification of the dictionary to reduce the error of activated words: new (warm) words are added to the dictionary to account for unparsed portions of the utterance, and variations of words used in the parse are added to the dictionary to fix mismatched or overparsed[4] portions of the utterance. Periodically words are deleted from the dictionary if they have not been cooled by being used, a brand of prior that favors a minimal-size dictionary.

---

PROCESS-UTTERANCE($u$, $d$) = {
        let $words$ = DICTIONARY-WORDS($d$)
(§3.1)     let $matches$ = MATCH-WORDS($u$, $words$)
(§3.2)     let $E^1$, $<\alpha^{1w}>$, $<\Delta P_i^1>$, $<\Delta S_j^1>$ = PARSE($u$, $matches$)
(§3.3)     let $new\text{-}words$ = CREATE-NEW-WORDS($u$, $matches$, $<\alpha^{1w}>$, $<\Delta P_i^1>$, $<\Delta S_j^1>$)
(§3.1)     let $new\text{-}matches$ = MATCH-WORDS($u$, $new\text{-}words$)
(§3.2)     let $E^2$, $<\alpha^{2w}>$, $<\Delta P_i^2>$, $<\Delta S_j^2>$ = PARSE($u$, $matches$+$new\text{-}matches$)
(§3.4)     COOL-WORDS($matches$+$new\text{-}matches$, $E^2$, $<\alpha^{2w}>$, $<\Delta P_i^2>$)
(§3.4)     ADD-COOLED-WORDS($new\text{-}matches$, $d$)
(§3.5)     GARBAGE-COLLECT-DICTIONARY($d$)
(§3.5)     REDUCE-DICTIONARY($d$)
}

Figure 1: Pseudocode for the procedure the learning algorithm applies to each utterance. $u$ is an utterance, $d$ is the dictionary, $E$ is an error scalar, $<\alpha^w>$ is a vector of word activations, $<\Delta P_i>$ is a vector of phonemic deviances, and $<\Delta S_j>$ is a vector of semantic deviances. Subroutines are described below in the sections listed.

---

Figure 1 presents pseudocode for the algorithm. The subroutines used by this algorithm are described in more detail below.

## 3.1 Matching

A word may occur at different points in a phonemic utterance. The MATCH-WORDS function of the algorithm finds places in an utterance that a word might occur and generates an evaluation of how closely it matches there. It does this by creating a vector (the *phonemic-match*, or $PM$ vector) that describes in terms of numbers from 0 to 1[5] how well a word accounts for each phoneme in the utterance, and a similar $SM$ vector for the sememes. It also computes two scalars, $P\overline{M}$ and $S\overline{M}$ that represent mismatched phonemes and sememes (such as a sememe in the word but not in the utterance).

---

[4] By *overparsed* we mean portions of the utterance that are accounted for by more than one word.

[5] Right now only 0 or 1 are used, 1 for a perfect match and 0 for anything else. This scheme anticipates using intermediate values to represent phonemes likely to be related by phonological processes, such as /t/ and /d/.

| Word | | Position | $PM$ /ðəmɛn/ | $SM$ { THE MAN } | $P\overline{M}$ | $S\overline{M}$ |
|---|---|---|---|---|---|---|
| /ðə/ | { THE } | 0 | < 11000 > | < 10 > | 0 | 0 |
| /ðə/ | { THE } | 1 | < 00000 > | < 10 > | 2 | 0 |
| /ðɛm/ | { THEM } | 0 | < 10100 > | < 00 > | 1 | 1 |
| /man/ | { MAN } | 2 | < 00101 > | < 01 > | 1 | 0 |
| ⋮ | ⋮ | ⋮ | ⋮ | ⋮ | ⋮ | ⋮ |

Figure 2: Some of the data produced by MATCH-WORDS to evaluate how well the words *the*, *them*, and *man* match up with the utterance *the men*. The position reflects the offset of the phonemic word into the utterance. Thus the word /ðə/ matches well phonetically when offset 0 and poorly when offset 1.

Figure 2 contains a sample of what the MATCH-WORDS function produces for the utterance *the men*, assuming the dictionary contains reasonable definitions for the words *the*, *them* and *man*. The match with the word *the* offset into the utterance by 1 has a sufficiently poor phonemic match that MATCH-WORDS would filter it out. The function returns a list of these matches and the associated vectors and scalars.

## 3.2 Parsing

In order to evaluate how well the current dictionary accounts for an utterance, the algorithm attempts to fully parse the utterance with words from the dictionary, placing words in such a fashion that each phoneme of the utterance is covered by a phoneme from exactly one dictionary word (with no extra phonemes being contributed by any word), and each sememe from the utterance is covered by a sememe from exactly one dictionary word (with no extra sememes being contributed by any word).

These desiderata can be modeled by giving each word match $w$ an activation coefficient $\alpha^w$ between 0 and 1. If $\alpha^w = 0$, the word does not participate in the parse. If $\alpha^w = 1$, the word participates fully. A perfect parse meets the following conditions using non-fractional activations:

$$\sum_w \alpha^w PM^w = <1\ldots 1>, \quad \sum_w \alpha^w SM^w = <1\ldots 1>$$

$$\sum_w \alpha^w P\overline{M}^w = 0, \quad \sum_w \alpha^w S\overline{M}^w = 0$$

The first two conditions guarantee that every phoneme and sememe is covered exactly once by the words in the utterance.[6] The second two guarantee that these words are not also contributing any extraneous phonemes or sememes. Of course, it may not be possible to meet these conditions given the available dictionary, at least not without fractional activations. So we parse with the goal of minimizing a global error function. If we let $\Delta P_i = 1 - \sum \alpha^w PM_i^w$, the distance between the current total activation of the $i$th phoneme and its target of 1, and similarly for sememes $\Delta S_j$, our global error function is

$$E = c_1 \sum_i f(\Delta P_i) + c_2 \sum_j f(\Delta S_j) + c_3 \sum_w \alpha^w P\overline{M}^w + c_4 \sum_w \alpha^w S\overline{M}^w$$

---
[6]Actually, the semantic target is not necessarily a uniform 1 vector, since some sememes may occur multiple times in the utterance. One might alternatively leave the target vector at a uniform 1 and adjust the success requirement to $\sum \alpha_i^w SM_i^w \geq 1$. A pause can be represented with a 0 phonemic target.

Here $i$ sums over the length of the phonemic utterance, $j$ over the number of different sememes in the utterance, and $w$ over the word matches. The $f$ function must be carefully chosen to result in a concentration of error in single phonemes or sememes rather than a distribution over the parse, and to penalize over-parsing a phoneme or sememe.[7]

We can minimize $E$ by varying the activation vector $<\alpha^w>$. A simple gradient-descent search from a randomly placed starting vector performs adequately for the particular vectors that arise here. The end result of the parsing process is a tuple of the final minimized error $E$, the activation vector $<\alpha^w>$, and the deviation vectors $<\Delta P_i>$ and $<\Delta S_i>$. Thus at the end of the parsing process we know not only how much each word participates in the parse ($\alpha^w$) but also which phonemes and sememes are under or over-parsed ($\Delta P_i$ and $\Delta S_j$). In the terms of the EM framework we now have an estimate of the hidden variables: the word activations.

## 3.3 Creating New Words

Once a parse of an utterance has been completed, the algorithm has some sense of what words participated in the utterance and what was misparsed; it now can perform the maximization step of modifying the dictionary. It creates new words, using a variety of methods that have proven successful but are not in any way the only ones that might work. Some of the methods used in this process are similar to those used by Siskind [13]. We can divide the methods into two parts: fixing words that participated in the parse and creating wholly new words. Fixes include deleting and adding phonemes and sememes from a definition.

Words participate in fixes with some probability. In the case of semantic fixes, that probability is proportional to the word's activation and its temperature. This prevents a cold word from participating in any semantic changes. In the case of phonemic fixes, the probability is proportional only to the word's activation. This makes it easy for a fully frozen word (say *cucumbers*) to create a new word *cucumber* with the same meaning, but difficult for *cucumbers* to change its meaning to { AVOCADO }. The fixes that a word can participate in are:

- Remove sememes from the word if they do not occur in the utterance sememe set or are overparsed ($\Delta S_i < -c$).

---

[7]The following function performs adequately:

$$f(\delta) = \begin{cases} |\delta| + \epsilon(1 - 4(|\delta| - \frac{1}{2})^2) & \text{if } |\delta| \leq 1 \\ \delta^2 & \text{otherwise} \end{cases}$$

It penalizes error and makes it least expensive to concentrate error on some phonemes and sememes rather than to distribute error with partial activations.

- Add underparsed sememes ($\Delta S_i > c$) to the word if there are no underparsed phoneme sequences in the utterance (as then the misparse would most likely be due to a missing word).

- Alter the word's phoneme sequence so as to eliminate phonemes that mismatch with the utterance, and to eliminate phonemes that are overparsed ($\Delta P_i < -c$).

- Extend the word's phoneme sequence so as to incorporate neighboring underparsed phonemes ($\Delta P_i > c$), up to a certain maximal length of extension.

In all cases the original word remains in the dictionary and a new word is created that incorporates the change.

Wholly new words are also created to account for unparsed portions of the utterance. A set of sequences of consecutive underparsed phonemes from the original utterance is created. These sequences represent the phonemic components of potential new words. Similarly, a set of the underparsed sememes is created. If there are two or fewer underparsed phoneme sequences and each is below a maximum new-word-length, then each one is turned into a new word, using the set of underparsed sememes as the hypothesized semantic representation.

New words start out with a high temperature, near 1.

## 3.4 Cooling Words

As can be seen from the pseudocode in figure 1, new words are used in a reparse of the utterance. If the result is a good parse and these words are highly activated, then confidence in the words is increased by cooling the temperature asymptotically towards 0. The COOL-WORDS subroutine of the algorithm cools a word from a parse if it meets each of several conditions:

- It has no phonemic mismatches ($P\overline{M}^w = 0$).

- It has no semantic mismatches ($S\overline{M}^w = 0$).

- Its neighboring phonemes are well parsed ($\Delta P_{l,r} < c$, where $l$ and $r$ are the left and right phonemic boundaries of the word match).

- Its activation is over a threshold ($\alpha^w > c$)

Cooling is a function of the total parse error $E$. A low error implies more cooling. Words are therefore cooled when they are confidently used in a successful parse. A nearly-frozen word has successfully taken part in a number of good parses. A warm word has not reliably demonstrated its necessity.

New words are not added to the dictionary unless they are cooled after the reparse of the utterance that caused their creation. This minimizes the number of potentially-disruptive changes to the dictionary.

## 3.5 Removing Words from the Dictionary

As utterances are parsed, new words are created to explain and correct errors, and are added to the dictionary. Many of these new words are unsuccessful and do not participate in many parses; they represent failed branches of the search. If a word remains uncooled for some time period, it is a good indication that adding that word to the dictionary was a mistake. After a certain fixed-length trial number of utterances, a word becomes open for deletion from the dictionary. Periodically words are garbage-collected from the dictionary, with the probability of deletion roughly proportional to the temperature. A fully frozen word (temperature = 0) will never be deleted. A warm word is highly likely to be deleted.[8]

As the algorithm starts to learn with an empty dictionary, the first words it creates tend to be utterance-encompassing, such as /atsraɪt/ { THAT BE RIGHT }. Later the algorithm learns the components of such words, i.e., /at/, /s/, and /raɪt/. Periodically the dictionary attempts to reduce its size by parsing each of its words. If it can successfully parse a word without recourse to the word itself, that word is eliminated from the dictionary.

The process of removing words from the dictionary is a means of implementing a prior preference for a small dictionary, one with no unnecessary words. The gradual cooling of words used in parses ensures that words remain in the dictionary only if the data necessitate their presence.

## 4 A Short Example

Here we present a short description of the algorithm's performance on a single example from the test suite, /yukɪktɔffðəsɑk/ paired with { YOU KICK OFF THE SOCK }. At the point that the utterance is encountered, the matching process finds 3 acceptably close matches in the dictionary: /yu/ { YOU } offset 0, /ðə/ { THE } offset 9, and /rsɑk/ { SOCK } offset 10. Notice that *you* and *the* have no mismatches, but *sock* has an extra /r/. *You* and *the* are well cooled (temperature near 0) at this point, but not surprisingly, *sock* is still quite warm (temperature .64).

Parsing with these three words leaves all with activation near 1. Two new words are then created. The first is a fix of the phonemic mismatch in /rsɑk/. It is /sɑk/ { SOCK }, the old word with the one mismatched phoneme removed. The second word is completely new, created to account for the unparsed parts of the utterance: /kɪktɔff/ { KICK OFF }. The sentence is then rematched and parsed. In the new parse, /rsɑk/ is given low activation because the sentence can be parsed with less error using /sɑk/ instead, and /kɪktɔff/ is given activation near 1. The total error is quite low (it would be zero if the gradient descent search had produced correct activations of exactly 0 or 1), and the activated words are cooled. Thus, /sɑk/ and /kɪktɔff/ are cooled but /rsɑk/ is not. Garbage collection will eventually remove /rsɑk/ from the dictionary, which is not likely to be cooled again given the new competition from /sɑk/.

---

[8]This heuristic prevents the algorithm from learning words that only occur once or twice, a problem given Carey's [3] evidence that children can (and need to) acquire some words from a very small number of exposures. One solution would be to speed the cooling process as the majority of the dictionary becomes stable. But the problem has deep roots and needs greater investigation: any on-line algorithm that maintains no explicit memory of previous data points will have a difficult time recovering from some of its mistakes. The usual solution of weight-decay towards a prior (to improve generalization and allow an algorithm to recover from noisy or misinterpreted data) does not work well if the algorithm must maintain perfect memory.

## 5 Tests and Results

To test the algorithm, we are using 34438 utterances from the Childes database of mothers' speech to children [8],[14]. These text utterances were run through a publicly available text-to-phoneme engine and also used to create a semantic dictionary, in which each root word from the utterances was mapped to a corresponding sememe. Various forms of a root ("see", "saw", "seeing") all map to the same sememe, *e.g.*, SEE . Semantic representations for a given utterance are merely unordered bags of sememes generated by taking the union of the sememe for each word in the utterance. Figure 3 contains the first 6 utterances from the database.

| Sentence | Phoneme Sequence | Sememe Set |
| --- | --- | --- |
| this is a book. | /ðɪsɪzəbuk/ | { THIS BE A BOOK } |
| what do you see in the book? | /wɑtduyusiɪnðəbuk/ | { WHAT DO YOU SEE IN THE BOOK } |
| how many rabbits? | /haʊmɛnirabbɪts/ | { HOW MANY RABBIT } |
| how many? | /haʊmɛni/ | { HOW MANY } |
| one rabbit. | /wʌnrabbɪt/ | { ONE RABBIT } |
| what is the rabbit doing? | /wɑtɪzðərabbɪtduɪŋ/ | { WHAT BE THE RABBIT DO } |

Figure 3: The first 6 utterances from the Childes database used to test the algorithm.

We describe the results of a single run of the algorithm, trained on one exposure to each of the 34438 utterances. Successive runs tend to result in nearly identical dictionaries. The final dictionary contains 1182 words (some entries are different forms of a common stem). Over the corpus the algorithm has been exposed to 2158 different stems. 82 of the words in the dictionary have never been used in a low-error parse. We eliminate these words, most of which are high temperature, leaving 1100. Figure 4 presents some entries in the final dictionary, and figure 5 presents all 21 of the 1100 entries that could be considered significant mistakes. So 1079 out of the 1100 entries (98%) are correct.

The most obvious error visible in figure 5 is the suffix *-ing* (/ɪŋ/), which should be semanticless (have an empty sememe set). Indeed, a semanticless word is properly hypothesized but a special mechanism prevents semanticless words from being added to the dictionary. This mechanism is necessary because the error function overpromotes semanticless words and results in poor learning of phonological words that happen to contain them as substrings. Without it, the system would chance upon a new word like *ring*, /rɪŋ/, use the semanticless /ɪŋ/ to account for most of the sound, and build a new word /r/ { RING } to cover the rest; witness *something* in figure 5. One solution to such a problem is to incorporate additional linguistic knowledge about word structure and about sound changes that occur at word boundaries[9], a solution discussed to some extent by Cartwright and Brent [4] and Church [5]. Alternatively, as a more immediate workaround, we could provide the test corpus with an explicit semantic clue, such as the following association: /ɪŋ/ { PROGRESSIVE }.

---
[9]For instance, in English no stem may be vowel-less, and word boundaries can sometimes be dis-

| Phoneme Sequence | Sememe Set | Phoneme Sequence | Sememe Set |
|---|---|---|---|
| /yu/ | { YOU } | /bik/ | { BEAK } |
| /ðə/ / | { THE } | /we/ | { WAY } |
| /wɑt/ | { WHAT } | /bukkes/ | { BOOKCASE } |
| /tu/ | { TO } | /brik/ | { BREAK } |
| /du/ | { DO } | /fɪŋgɜ/ | { FINGER } |
| /e/ | { A } | /santəklɔs/ | { SANTA CLAUS } |
| /ɪt/ | { IT } | /tɑp/ | { TOP } |
| /aɪ/ | { I } | /kɔld/ | { CALL } |
| /ɪn/ | { IN } | /ɛgz/ | { EGG } |
| /wi/ | { WE } | /θɪŋ/ | { THING } |
| /s/ | { BE } | /kɪs/ | { KISS } |
| /ɑn/ | { ON } | /hi/ | { HEY } |

Figure 4: Dictionary entries. The left 12 are the 12 words used most frequently in good parses. The right 12 were selected randomly from the 1100 entries.

Most other semanticless affixes (plural /s/ for instance) are also properly hypothesized and disallowed, but the dictionary learns multiple entries to account for them (/g/ "egg" and /gz/ "eggs"). The system learns synonyms ("is", "was", "am", ...) and homonyms ("read", "red"; "know", "no") without difficulty.

## 6 Shortcomings and Future Directions

The most obvious immediate shortcoming of the algorithm is the previously discussed difficulty with semanticless words and affixes. As mentioned in the introduction, a more important issue is the simplicity of the environment the current algorithm assumes. We are building a new and considerably more complex learning architecture to rectify some of the deficiencies. In particular,

- instead of discrete phonemes, the new system accepts a time series of potentially noisy estimated vocal articulator positions. It attempts to find phoneme sequences from its dictionary that provide the most complete and consistent account for the perceived positions.

- the system incorporates some knowledge of morphology and phonology, selected to enable it to infer some contextual sound-change rules and consequently use sound changes as evidence of word boundaries. This knowledge should help the system learn semanticless words and affixes.

---

tinguished with the knowledge that word-initial obstruents are aspirated (/t/ is pronounced with an exhalation of air in *top* but not *stop*).

| Phoneme Sequence | Sememe Set | Phoneme Sequence | Sememe Set |
|---|---|---|---|
| /ɪŋ/ | { BE } | /nupis/ | { SNOOPY } |
| /ɪŋ/ | { YOU } | /wo/ | { WILL } |
| /ɪŋ/ | { DO } | /zu/ | { AT ZOO } |
| /ʃiz/ | { SHE BE } | /don/ | { DO } |
| /wɑthappind/ | { WHAT HAPPEN } | /ɛrɛ/ | { BE } |
| /dont/ | { DO NOT } | /smʌd/ | { MUD } |
| /sʌmθ/ | { SOMETHING } | /nidəlɪz/ | { NEEDLE BE } |
| /wɑtɑrðiz/ | { WHAT BE THESE } | /drɑnʌðɜwiz/ | { DROWN OTHERWISE } |
| /shappin/ | { HAPPEN } | /sɛlf/ | { YOU } |
| /t/ | { NOT } | /ə/ | { BE } |
| /skɑtt/ | { BOB SCOTT } | | |

Figure 5: All of the significant dictionary errors. Some of them, like /ʃiz/ are conglomerations that should have been divided. Others, like /t/, /wo/, and /don/ demonstrate how the system compensates for the morphological irregularity of English contractions. The /ɪŋ/ problem is discussed in the text; misanalysis of the role of /ɪŋ/ also manifests itself on *something*.

- the system accepts a complex conditional probability distribution over semantic symbols as the child's interpretation of the environment, and uses a Bayesian model to determine which sememes are actually represented by the utterance.

If successful, the improved system will considerably expand the scope and performance of the preliminary work presented here.

# 7 Acknowledgements

This research is supported by NSF grant 9217041-ASC and ARPA under the HPCC program. David Baggett, Robert Berwick, Jeffrey Siskind, Oded Maron, Greg Galperin and Marina Meila have made valuable suggestions and comments related to this work. A Common Lisp version of the algorithm and the test corpus can be found in `ftp://ftp.ai.mit.edu/pub/users/cgdemarc/papers/icgi94`.